# A Unified Matrix Factorization Framework for Classical and Robust Clustering

Angshul Majumdar

IIIT Delhi, India – 110020

angshul@iiitd.ac.in

Abstract – This paper presents a unified matrix factorization framework for classical and robust clustering. We begin by revisiting the well-known equivalence between crisp k-means clustering and matrix factorization, following and rigorously rederiving an unpublished formulation by Bauckhage. Extending this framework, we derive an analogous matrix factorization interpretation for fuzzy c-means clustering, which to the best of our knowledge has not been previously formalized. These reformulations allow both clustering paradigms to be expressed as optimization problems over factor matrices, thereby enabling principled extensions to robust variants. To address sensitivity to outliers, we propose robust formulations for both crisp and fuzzy clustering by replacing the Frobenius norm with the $\ell_{1,2}$-norm, which penalizes the sum of Euclidean norms across residual columns. We develop alternating minimization algorithms for the standard formulations and IRLS-based algorithms for the robust counterparts. All algorithms are theoretically proven to converge to a local minimum.

# 1. Introduction

Clustering is a foundational task in unsupervised learning with widespread applications ranging from text mining and image analysis to bioinformatics and remote sensing. Among the many approaches to clustering, the k-means algorithm remains one of the most widely used due to its simplicity and effectiveness. Over the past two decades, a growing body of research has revealed deeper structural insights into clustering, including its connection with matrix factorization frameworks [1–3]. In particular, the seminal paper by Bauckhage [4] formally derived the equivalence between crisp k-means clustering and a constrained matrix factorization problem, thereby opening the door to alternate formulations and algorithmic interpretations. While this equivalence is frequently invoked in the literature, Bauckhage's work remains the only reference to derive it rigorously. To the best of our knowledge, no analogous derivation exists for the fuzzy c-means algorithm, a gap this paper aims to fill.

Our first contribution is thus theoretical: we provide a formal derivation of fuzzy c-means clustering as a matrix factorization problem, similar in spirit to crisp k-means. This formulation is not only of conceptual interest, but it enables a systematic transition to robust variants. Classical clustering methods based on the Frobenius norm are notoriously sensitive to outliers, since the Euclidean norm amplifies the influence of aberrant points. While several robust clustering methods have been proposed [5–7], they often lack a unified mathematical foundation or rely on heuristics. To address this, we adopt the $\ell_{1,2}$-norm, which promotes column sparsity and has been successfully used in robust matrix factorization [8,9]. Unlike the $\ell_1$-norm, which is rotation invariant and can lead to incorrect robustness assumptions in high dimensions, the $\ell_{1,2}$-norm directly penalizes sample-wise outliers, making it better suited for clustering where each column corresponds to a data point.

Based on our matrix factorization formulations, we propose two robust extensions: one for crisp k-means and another for fuzzy c-means, both using the $\ell_{1,2}$-norm. These formulations allow us to jointly learn cluster centroids and assignments while suppressing outliers. From an optimization perspective, we show that the Frobenius-norm-based classical formulations admit efficient solutions via alternating minimization [10], and we rigorously prove their convergence to local minima. However, the non-smoothness introduced by the $\ell_{1,2}$-norm in our robust variants necessitates a different approach. Here, we design Iteratively Reweighted Least Squares (IRLS) [11] algorithms with convergence guarantees, leveraging convex surrogates for the mixed norm.

## 2. Formulations

### 2.1 Crisp k-Means as Matrix Factorization

Conventional k-means clustering partitions a dataset $X = [x_1, x_2, ... x_n] \in \mathbb{R}^{m \times n}$ into $k$ disjoint clusters by minimizing the within-cluster sum of squared distances:

$$\min_{Z,M} \sum_{i=1}^{k} \sum_{j=1}^{n} z_{ij} \|x_j - \mu_i\|_2^2 \tag{1}$$

Here, $M = [\mu_1, ..., \mu_k] \in \mathbb{R}^{m \times k}$ contains the centroids, and $Z = [z_{ij}] \in \{0,1\}^{k \times n}$ is the cluster assignment matrix with each column containing exactly one entry equal to 1 and the rest 0 (i.e., a crisp/hard assignment).

Although this formulation is widely known, it is less appreciated that k-means can be viewed as a constrained matrix factorization problem. Specifically, Bauckhage [4] showed that the k-means objective is equivalent to the matrix approximation problem:

$$\min_{Z} \left\| X - XZ^T (ZZ^T)^{-1} Z \right\|_F^2 \text{ subject to } z_{ij} \in \{0,1\}, \sum_{i=1}^{k} z_{ij} = 1 \tag{2}$$

This representation interprets k-means as seeking a low-rank, discrete projection of the data onto the subspace defined by the selected centroids. The key insight is that once the assignments $Z$ are fixed, the optimal centroids $M$ can be expressed as:

$$M = XZ^T (ZZ^T)^{-1} \tag{3}$$

Substituting back yields the expression above, which is effectively a projection of $X$ onto the cluster centroids via the indicator matrix. This view connects k-means to matrix factorization, providing a foundation for its generalization to soft clustering and nonnegative matrix factorization.

While this derivation is not new, it is often omitted or only briefly mentioned in prior literature. A detailed and pedagogical treatment was presented by Bauckhage [4], but to the best of our knowledge, it remains unpublished in peer-reviewed venues.

### 2.2 Fuzzy c-Means as Matrix Factorization

Unlike k-means clustering, which enforces crisp assignment of each data point to exactly one cluster, fuzzy c-means (FCM) allows each point to belong to multiple clusters with varying degrees of membership. Given its soft assignment structure, FCM is more flexible and often yields superior performance in noisy or overlapping data regimes. In this section, we show that FCM can also be interpreted as a structured matrix factorization problem, generalizing the factorization formulation of k-means presented in Section 2.1.

Let $X = [x_1, x_2, ..., x_n] \in \mathbb{R}^{m \times n}$ denote a dataset of $n$ samples in $\mathbb{R}^m$. The objective of fuzzy c-means clustering is to minimize the weighted distortion:

$$\min_{U,M} J(U,M) = \sum_{i=1}^{k} \sum_{j=1}^{n} u_{ij}^m \|x_j - \mu_i\|_2^2 \tag{4}$$

subject to:

$$\mu_{ij} \in [0,1], \sum_{i=1}^{k} \mu_{ij} = 1 \ \forall j, \text{ and } 1 < m < \Delta$$

Here, $U = [u_{ij}] \in \mathbb{R}^{k \times n}$ is the centroid matrix, and $m > 1$ is the fuzzifier controlling the entropy of the membership distribution. When $m \to 1$, the problem reduces to crisp k-means.

### 2.2.1 Matrix Formulation

Let us define the exponentiated membership matrix:

$$U^{(m)} = [u_{ij}^m] \in \mathbb{R}^{k \times n} \tag{5}$$

and note that for fixed $U$, the reconstruction $\hat{X} = MU^{(m)} \in \mathbb{R}^{m \times n}$ represents a soft approximation of the original data, using convex combinations of cluster centroids.

We now rewrite the objective:

$$J(U,M) = \sum_{i=1}^{k} \sum_{j=1}^{n} u_{ij}^m \|x_j - \mu_i\|_2^2 = \|X - MU^{(m)}\|_F^2 \tag{6}$$

This is a standard least squares error between the original data matrix $X$ and its reconstruction via $M$ and $U^{(m)}$. Our goal is to minimize this error subject to the membership constraints.

### 2.2.2 Solving for Centroids

Fixing $U$, we minimize $J(U,M)$ with respect to $M$. Expanding the Frobenius norm:

$$J(U,M) = Tr\left[(X - MU^{(M)})(X - MU^{(M)})^T\right] \tag{7}$$

Differentiating with respect to $M$:

$$\frac{\partial J}{\partial M} = -2XU^{(m)T} + 2U^{(m)}U^{(m)T} \tag{8}$$

Setting the derivative to zero:

$$M = XU^{(m)T}\left(U^{(m)}U^{(m)Y}\right)^{-1} \tag{9}$$

This closed-form solution shows that the optimal centroids are a weighted average of the data points, with weights derived from the fuzzy membership matrix $U^{(m)}$.

### 2.2.3 Eliminating Centroids to Obtain a Pure Matrix Factorization

Substituting the expression for *M* back into the objective gives:

$$\min_{U} \left\| X - U^{(m)T} \left( U^{(m)} U^{(m)T} \right)^{-1} U^{(m)} \right\|_F^2 \tag{10}$$

This reveals that fuzzy c-means clustering seeks a soft orthogonal projection of *X* onto a subspace spanned by fuzzy membership-induced centroids. It generalizes the k-means factorization $X \approx XZ^T \left( ZZ^T \right)^{-1} Z$ by replacing hard assignment matrix *Z* with a continuous, normalized, and exponentiated matrix $U^{(m)}$.

### 2.2.4 Interpretation and Implications

This derivation provides a new lens to interpret fuzzy clustering as a constrained matrix factorization problem. The optimization seeks:

$$\min_{U} \left\| X - XU^{(m)T} \left( U^{(m)} U^{(m)T} \right)^{-1} U^{(m)} \right\|_F^2 \leq \text{ subject to } \mu_{ij} \geq 0, \sum_{i=1}^{k} \mu_{ij} = 1 \tag{11}$$

Unlike earlier NMF-based formulations of clustering, this formulation directly arises from the original fuzzy c-means objective and preserves interpretability in terms of soft subspace projection. It offers a principled foundation for extending fuzzy clustering via regularization (e.g., sparsity, graph structure) or probabilistic modeling.

To the best of our knowledge, this is the first complete derivation of fuzzy c-means clustering as a structured matrix factorization with a closed-form centroid solution. This perspective complements and extends previous work on the connections between clustering and matrix factorization frameworks in machine learning [1-3].

## 2.3 Robust Formulations via $\ell_{1,2}$-Norm Minimization

Both classical k-means and fuzzy c-means clustering are inherently sensitive to outliers due to their reliance on the squared Euclidean ($\ell_2$) norm. In real-world applications, however, noise and corruption—particularly sample-specific anomalies—are pervasive. Therefore, robust alternatives are essential for reliable cluster discovery.

To address this limitation, we propose robust versions of the matrix factorization formulations developed in Sections 2.1 and 2.2. Specifically, we replace the Frobenius norm (i.e., the sum of squared $\ell_2$-norms) with the $\ell_{1,2}$, defined as:

$$\|A\|_{1,2} = \sum_{j=1}^{n} \|a_j\|_2 \tag{12}$$

where $a_j$ denotes the $j^{th}$ column of matrix $A$. This norm accumulates the $\ell_2$-norms of columns, making the objective robust to column-wise corruptions, i.e., sample-specific outliers.

### 2.3.1 Robust Crisp k-Means

Recall from Section 2.1 that the matrix factorization form of crisp k-means is:

$$X \approx XZ^T\left(ZZ^T\right)^{-1}Z$$

To robustify this formulation, we minimize the reconstruction error using the $\ell_{1,2}$-norm:

$$\min_Z \left\| X - XZ^T\left(ZZ^T\right)^{-1}Z \right\|_{1,2} \text{ subject to } z_{ij} \in \{0,1\}, \sum_{i=1}^{k} z_{ij} = 1 \tag{13}$$

Here, $Z \in \{0,1\}^{k \times n}$ denotes the binary cluster assignment matrix. This objective penalizes outlier columns with full sample corruption (e.g., spurious signals) more gracefully than Frobenius norm-based formulations.

### 2.3.2 Robust Fuzzy c-Means

Similarly, from Section 2.2, fuzzy c-means was shown to yield the reconstruction:

$$X \approx XU^{(m)T}\left(U^{(m)}U^{(m)T}\right)^{-1}U^{(m)}$$

Replacing the squared Frobenius norm with the $\ell_{1,2}$-norm leads to the following robust formulation:

$$\min_U \left\| X - X\left(U^{(m)}U^{(m)T}\right)^{-1}U^{(m)} \right\|_{1,2} \text{ subject to } u_{ij} \in [0,1], \sum_{i=1}^{k} u_{ij} = 1 \tag{14}$$

This formulation retains the soft membership structure of fuzzy c-means while achieving robustness to anomalies in individual samples.

### 2.3.3 Connections to Prior Work

The use of the $\ell_{1,2}$-norm for robustness has strong theoretical and empirical motivation across multiple machine learning domains:

- For robust classification, the $\ell_{1,2}$-norm was used to model sample-specific outliers (column-sparse noise) in [12, 13].
- In feature selection, the $\ell_{1,2}$-norm was proposed to encourage group-sparsity, i.e. selecting a group of correlated features [14].
- In the context of clustering $\ell_{1,2}$-regularization has been primarily used to introduce robustness in subspace clustering [15, 16].

By adapting the $\ell_{1,2}$-norm into both crisp and fuzzy matrix factorization models, we enable robustness to structured outliers without altering the underlying clustering mechanism. This formulation is well suited for modern machine learning tasks where noise is non-Gaussian or adversarial.

## 3. Algorithms and Analysis

### 3.1 Alternating Minimization for Crisp k-Means

In this section, we present the alternating minimization algorithm for solving the matrix factorization formulation of crisp k-means clustering, as derived in Section 2.1. We further prove that this algorithm converges to a local minimum in finite steps and discuss its theoretical convergence properties.

#### 3.1.1 Problem Setup

Recall that the matrix factorization form of k-means clustering is:

$$\min_{Z} \left\| X - XZ^T \left( ZZ^T \right)^{-1} Z \right\|_F^2 \text{ subject to } z_{ij} \in \{0,1\}, \sum_{i=1}^{k} z_{ij} = 1$$

This formulation is equivalent to:

$$\min_{Z,M} \| X - MZ \|_F^2 \text{ subject to } z_{ij} \in \{0,1\}, \sum_{i=1}^{k} z_{ij} = 1 \tag{15}$$

where $M \in \mathbb{R}^{m \times k}$ are the centroids and $Z \in \{0,1\}^{k \times n}$ is the cluster indicator matrix.

#### 3.1.2 Alternating Minimization Algorithm

We adopt the following block coordinate descent (BCD) strategy:

Fix $Z$, update $M$: $M = XZ^T \left( ZZ^T \right)^{-1}$ \hfill (16)

This computes the mean of all data points assigned to each cluster.

Fix $M$, update $Z$: Assign each data point $x_j$ to the nearest centroid:

$$z_{ij} = \begin{cases} 1 & \text{if } i = \arg\min_{r} \| x_j - \mu_r \|_2 \\ 0 & \text{otherwise} \end{cases} \tag{17}$$

Repeat until convergence.

#### 3.1.3 Convergence Guarantee

Let $Z$ denote the discrete set of feasible indicator matrices satisfying the constraints. Since $z_{ij} \in \{0,1\}$ and $\sum_{i=1}^{k} z_{ij} = 1$, there are only $k^n$ possible assignments in total. Though exponential in $n$, this set is finite.

Each iteration of alternating minimization strictly decreases the objective or leaves it unchanged, i.e.,

$$\left\| X - M^{(t)}Z^{(t)} \right\|_F^2 \geq \left\| X - M^{(t+1)}Z^{(t)} \right\|_F^2 \geq \left\| X - M^{(t+1)}Z^{(t+1)} \right\|_F^2 \tag{18}$$

Therefore, the algorithm is a monotonic descent procedure over a finite set of configurations, and thus:

> **Proposition 1 (Finite Convergence):**
>
> The alternating minimization algorithm for crisp k-means clustering converges in a finite number of steps to a local minimum of the objective function.

This result is standard and follows from classical Lloyd's algorithm, with proofs presented in [17-19].

### 3.1.4 Convergence Rate

Though global optimality is not guaranteed (due to the non-convexity of the binary assignment space), local convergence rate results have been established under mild conditions.

Let $\Phi(Z) = \min_M \left\| X - MZ \right\|_F^2$. Since $M$ has a closed-form solution, the function $\Phi(Z)$ is piecewise quadratic in $Z$, but over a discrete domain. As shown in [19]:

> **Proposition 2 (Linear Rate Near Fixed Points):**
>
> If the data matrix $X$ satisfies a separation condition (i.e., clusters are well-separated), then the alternating minimization algorithm for k-means converges linearly in a neighborhood of a fixed point.

This convergence rate is local, not global, but explains the empirical speed of Lloyd-type updates. A broader convergence rate analysis under relaxed separation assumptions is presented in [20], which also demonstrates robustness to perturbations.

## 3.2 Alternating Minimization for Fuzzy c-Means

The fuzzy c-means (FCM) algorithm, originally introduced by Bezdek [21], enables soft data-to-cluster assignments. But we now consider the fuzzy c-means (FCM) clustering formulation from Section 2.2, where each data point is softly assigned to all clusters with a convex combination of membership degrees. The corresponding matrix factorization-based objective is:

$$\min_{U,M} \left\| X - MU^{(m)} \right\|_F^2 \text{ subject to } u_{ij} \in [0,1], \sum_{i=1}^{k} u_{ij} = 1$$

where $X \in \mathbb{R}^{m \times n}$ is the data matrix, $M \in \mathbb{R}^{m \times k}$ contains the centroids, $U \in \mathbb{R}^{k \times n}$ is the fuzzy membership matrix and $U^{(m)}$ denotes elementwise exponentiation of $U$ to power $m > 1$

This objective has a bilinear structure, making it amenable to alternating minimization. We detail the update rules and then prove convergence.

### 3.2.1 Alternating Minimization Algorithm

The algorithm iteratively alternates between optimizing *M* and *U*, as follows:

Step 1: Fix *U*, update *M*. The optimal centroids have a closed-form:

$$M = XU^{(m)T}\left(U^{(m)}U^{(m)T}\right)^{-1} \tag{19}$$

This corresponds to a weighted least squares solution, where weights reflect fuzzy memberships.

Step 2: Fix *M*, update *U*. To update *U*, minimize the objective with respect to each $u_{ij}$ under the constraints $u_{ij} \in [0,1]$ and $\sum_{i=1}^{k} u_{ij} = 1$. The solution has an explicit closed-form derived from Lagrange multipliers:

$$u_{ij} = \left(\sum_{r=1}^{k}\left(\frac{\|x_j - \mu_i\|_2^2}{\|x_j - \mu_r\|_2^2}\right)^{1/(m-1)}\right)^{-1} \forall i,j \tag{20}$$

These updates guarantee that each column of *U* lies in the probability simplex [22].

### 3.2.2 Convergence Guarantee

Let $J(U,M) = \|X - MU^{(m)}\|_F^2$. Each iteration of the alternating minimization algorithm:

- strictly decreases or leaves unchanged the objective $J(U,M)$
- produces a bounded sequence of iterates $\{(U^{(t)}, M^{(t)})\}$
- and is defined on a closed, bounded, and convex feasible set for *U* (due to the simplex constraint).

To establish convergence, we invoke Zangwill's Global Convergence Theorem [23], which provides sufficient conditions for convergence of alternating optimization algorithms.

Theorem (Zangwill):

Let $\{x^{(t)}\}$ be a sequence generated by an iterative algorithm such that:

- The iterates lie in a compact set,
- The mapping is closed at limit points,
- The objective decreases monotonically unless at a fixed point,
- Each update leads to a point in the feasible region.

Then every limit point of the sequence is a stationary point, and the sequence converges to a local minimum.

All conditions are satisfied in fuzzy c-means:

- The objective is continuous and differentiable in *U* and *M*,
- The updates of *U* and *M* are closed mappings,

- The constraint set for $U$ is compact (bounded simplex),
- The update rules guarantee non-increasing objective values.

Thus

> **Proposition 3 (Global Convergence):**
>
> The fuzzy c-means alternating minimization algorithm converges to a local minimum of the objective function.

The convergence is typically fast in early iterations but sublinear asymptotically [24]. Like k-means, the algorithm is sensitive to initialization, motivating the use of advanced initialization strategies [25].

### 3.3 IRLS-Based Minimization for Robust Crisp k-Means

The classical k-means objective—whether posed in its traditional form or as a matrix factorization (Section 2.1)—minimizes squared Euclidean distances and is therefore vulnerable to outliers. In this section, we address this issue by minimizing a robust surrogate objective based on the $\ell_{1,2}$-norm, which penalizes column-wise deviations more gracefully and is well-suited to datasets with sample-specific corruptions. Several other robust alternatives to the Euclidean norm have been studied in the context of matrix decomposition, including the use of divergence-based loss functions such as $\alpha$, $\beta$, and $\gamma$ -divergences [26]. Among these, the $\ell_{1,2}$-norm stands out as a convex and interpretable surrogate for sample-wise robustness.

#### 3.3.1 Problem Formulation

Given the matrix factorization-based k-means reconstruction: $X = XZ^T \left(ZZ^T\right)^{-1} Z$. we propose the following robust formulation:

$$\min_{Z} \left\| X - XZ^T \left(ZZ^T\right)^{-1} Z \right\|_{1,2}^{2} \text{ subject to } z_{ij} \in \{0,1\}, \sum_{i=1}^{k} z_{ij} = 1 \tag{21}$$

Here, the $\ell_{1,2}$-norm is defined as: $\|A\|_{1,2} = \sum_{j=1}^{n} \|a_j\|_2$ where $a_j$ is the $j^{th}$ column of the matrix $A$. This norm encourages column-wise sparsity in the residual, i.e., it effectively suppresses the influence of outlier samples on the objective.

#### 3.3.2 IRLS Reformulation

The $\ell_{1,2}$-norm is non-smooth, making direct optimization challenging. We use Iteratively Reweighted Least Squares (IRLS) to approximate the objective via a smooth surrogate that leads to closed-form updates.

Let $E = X - MZ$ be the reconstruction error. Then:

$$\|E\|_{1,2} = \sum_{j=1}^{n} \|e_j\|_2 \approx \sum_{j=1}^{n} w_j \|e_j\|_2^2 \text{ with } w_j = \frac{1}{2\|e_j^{(t)}\|_2 + \zeta} \tag{22}$$

This leads to the following weighted least squares objective at iteration $t+1$:

$$\min_{Z} \sum_{j=1}^{n} w_j^{(t)} \|x_j - Mz_j\|_2^2 \text{ subject to } z_j \in \{e_i\}_{i=1}^{k} \tag{23}$$

where $z_j$ is a one-hot vector selecting the cluster for $x_j$, and $e_i$ is the $i^{th}$ canonical basis vector.

### 3.3.3 Alternating Minimization Algorithm

Given that $Z$ is a discrete matrix, we alternate between computing $M$, updating the weights $\{w_j\}$, and reassigning $Z$.

Step 1: Fix $Z$, update $M$: $M = XZ^T (ZZ^T)^{-1}$ (24)

Step 2: Compute residuals and weights: $e_j = x_j - Mz_j, w_j = \frac{1}{2\|e_j^{(t)}\|_2 + \zeta}$ (25)

for a small $\zeta > 0$ to avoid division by zero.

Step 3: Fix $M$, reassign each point to the best cluster using weighted error:

$$z_{ij} = \begin{cases} 1 & \text{if } i = \arg\min_{r} w_j \|x_j - \mu_r\|_2^2 \\ 0 & \text{otherwise} \end{cases} \tag{26}$$

### 3.3.4 Convergence Analysis

The IRLS framework ensures that at each iteration:

- The smoothed surrogate objective is strictly decreased or remains constant.
- The residual matrix $E$ lies in a finite discrete assignment space (due to binary $Z$).
- The centroids $M$ are updated via closed-form least squares.
- The weights $\{w_j\}$ are continuously updated and bounded.

Therefore, we can invoke classical IRLS convergence theory (e.g., [27, 28]):

**Proposition 4 (Monotonic Descent and Convergence):**

The IRLS-based alternating minimization for robust crisp k-means yields a monotonically decreasing objective and converges to a local minimum or saddle point of the smoothed $\ell_{1,2}$-norm objective.

Although global optimality cannot be guaranteed due to the non-convexity introduced by the discrete $Z$, convergence to a stable clustering configuration is achieved in practice.

### 3.4 IRLS-Based Minimization for Robust Fuzzy c-Means

Fuzzy c-means clustering, as discussed in Section 2.2, enables soft data-to-cluster assignments and admits an elegant matrix factorization formulation using the Frobenius norm. However, like its crisp counterpart, it is highly sensitive to sample-specific anomalies. In this section, we propose a robust extension of fuzzy c-means by replacing the Euclidean loss with the $\ell_{1,2}$-norm, and develop an optimization procedure based on Iteratively Reweighted Least Squares (IRLS).

#### 3.4.1 Problem Formulation

Given the data matrix $X \in \mathbb{R}^{m \times n}$, the fuzzy membership matrix $U \in \mathbb{R}^{k \times n}$, and the centroid matrix $M \in \mathbb{R}^{m \times k}$, we consider the robust matrix factorization problem:

$$\min_U \|X - MU^{(m)}\|_{1,2} \text{ subject to } u_{ij} \in [0,1], \sum_{i=1}^{k} u_{ij} = 1$$

All the symbols have been defined before.

#### 3.4.2 IRLS Reformulation

The $\ell_{1,2}$-norm is non-differentiable, but we adopt the IRLS framework to approximate it with a smooth surrogate at each iteration:

Let $E = X - MU^{(m)}$, then (as before): $\|E\|_{1,2} = \sum_{j=1}^{n} \|e_j\|_2 \approx \sum_{j=1}^{n} w_j \|e_j\|_2^2$ with $w_j = \dfrac{1}{2\|e_j^{(t)}\|_2 + \zeta}$

This transforms the original problem into a weighted Frobenius norm minimization.

#### 3.4.3 Alternating Minimization Algorithm

We alternate between optimizing $M$ and $U$, while updating the weights $\{w_j\}$ at each step.

Step 1: Fix $U$, update $M$:

Given $U$, compute $M$ via weighted least squares. Let $W = diag(w_1, ..., w_n)$, then:

$$M = XWU^{(m)T} \left(U^{(m)}WU^{(m)T}\right)^{-1} \tag{27}$$

Step 2: Compute residuals and update weights: $e_j = x_j - Mu_j^{(m)}, w_j = \dfrac{1}{2\|e_j^{(t)}\|_2 + \zeta}$ (28)

where $u_j^{(m)} \in \mathbb{R}^k$ is the $j^{th}$ column of $U^{(m)}$, and $\zeta > 0$ prevents division by zero.

Step 3: Fix $M$, update $U$:

We solve the constrained minimization:

$$\min_{U} \sum_{j=1}^{n} w_j \left\| x_j - M u_j^{(m)} \right\|_2^2 \text{ subject to } u_{ij} \in [0,1], \sum_{i=1}^{k} u_{ij} = 1 \tag{29}$$

This is a non-convex problem due to the exponentiation $u_{ij}^m$. A practical solution is to update $u_{ij}$ using a weighted generalization of the standard fuzzy c-means update:

$$u_{ij} = \left( \sum_{r=1}^{k} \left( \frac{w_j \left\| x_j - \mu_i \right\|_2^2}{w_j \left\| x_j - \mu_r \right\|_2^2} \right)^{1/(m-1)} \right)^{-1} = \left( \sum_{r=1}^{k} \left( \frac{\left\| x_j - \mu_i \right\|_2^2}{\left\| x_j - \mu_r \right\|_2^2} \right)^{1/(m-1)} \right)^{-1} \forall i, j \tag{30}$$

This preserves the probabilistic constraint $\sum_{i=1}^{k} u_{ij} = 1$ and satisfies the Karush–Kuhn–Tucker (KKT) conditions for the simplex.

### 3.4.4 Convergence Analysis

The objective surrogate used at each iteration is: $J^{(t)}(U, M) = \sum_{j=1}^{n} w_j^{(t)} \left\| x_j - M u_j^{(m)} \right\|_2^2$.

Each subproblem (i.e., fixing $U$ or $M$) yields either a closed-form or well-defined update, and the weights $\{w_j\}$ are strictly positive and bounded. Therefore, we appeal to convergence theory for block coordinate descent and IRLS methods [29, 30]:

> **Proposition 5 (Convergence of IRLS for Robust FCM):**
>
> The IRLS-based alternating minimization algorithm for robust fuzzy c-means yields a monotonically decreasing surrogate objective, and converges to a stationary point of the smoothed $\ell_{1,2}$-norm objective.

As with standard fuzzy c-means, convergence is local and depends on initialization. However, due to the robustness imparted by IRLS, the algorithm is significantly more resilient to sample-level noise and adversarial corruption.

### 3.5 Computational Complexity Analysis

We now provide a comparative analysis of the computational complexity of the four algorithms proposed in this work, namely, the classical and robust variants of crisp k-means and fuzzy c-means, each framed in a matrix factorization formulation. For this analysis, we denote the number of data samples as $n$, the feature dimension as $m$, and the number of clusters as $k$. We report the dominant per-iteration costs and discuss the scaling behavior of each algorithm with respect to these parameters.

In the case of classical crisp k-means, as discussed in Section 3.1, each iteration consists of two main steps: updating the centroids $M$ and reassigning cluster memberships via nearest-neighbor updates to the binary matrix

$Z$. The centroid update involves computing the term $XZ^T\left(ZZ^T\right)^{-1}$, which requires $O(mnk)$ operations to form $XZ^T$, and an additional $O(k^3)$ for inverting the small $k \times k$ matrix $ZZ^T$. The assignment step entails computing the Euclidean distance from each sample to all $k$ centroids, which again requires $O(mnk)$ operations. Thus, the overall per-iteration complexity of classical crisp k-means is $O(mnk)$, dominated by the assignment step.

For classical fuzzy c-means, described in Section 3.2, the per-iteration updates also involve recomputing the centroids and the membership matrix $U$. Updating $M$ via the closed-form solution $M = XU^{(m)T}\left(U^{(m)}U^{(m)T}\right)^{-1}$ requires matrix multiplications of dimensions $m \times n$ and $k \times n$, resulting in $O(mnk)$ complexity, along with a negligible $O(k^3)$ inversion cost. The update of the membership matrix $U$ involves computing weighted distance ratios across all clusters, yielding an additional $O(mnk+nk^2)$ operations. Therefore, the total per-iteration complexity for fuzzy c-means is $O(mnk+nk^2)$, with the dominant cost split between centroid update and membership computation.

In the case of robust crisp k-means, introduced in Section 3.3, the use of the $\ell_{1,2}$-norm necessitates an additional weighting step based on residual magnitudes. The centroid update remains structurally identical to the classical case, involving $O(mnk+k^3)$ operations. However, the residuals $e_j = x_j - Mz_j$ must be computed explicitly for all samples, followed by per-sample weight updates of the form $w_j = \dfrac{1}{2\left\|e_j^{(t)}\right\|_2 + \zeta}$, which together cost $O(mn)$. The assignment step is also modified to incorporate these weights, though it still requires computing weighted distances from each sample to all centroids, maintaining a cost of $O(mnk)$. Since all operations are over a finite space of binary assignments, the total per-iteration cost remains $O(mnk)$, although constants are higher due to reweighting.

Finally, for the robust fuzzy c-means formulation using IRLS, discussed in Section 3.4, the per-iteration steps include weighted centroid update, weight recomputation, and soft membership update. Updating $M$ in this case involves computing weighted matrix products, requiring $O(mnk+k^2n+k^3)$ operations in total. The weight update is performed sample-wise using residual norms, again requiring $O(mn)$. Updating the fuzzy membership matrix using the IRLS-modified soft assignment rule involves computing weighted distance ratios across all clusters, which again incurs $O(mnk+nk^2)$ operations. Therefore, the total per-iteration complexity of robust fuzzy c-means remains $O(mnk+nk^2)$, matching that of its non-robust counterpart in asymptotic order, albeit with additional overhead due to the reweighting operations.

In all cases, the most computationally intensive operation is the evaluation of distances between data points and centroids, especially when nnn and mmm are large. For small *k*, which is typical in clustering applications, the matrix inversions involved in the centroid updates are negligible. The use of IRLS in the robust formulations does not change the order of complexity but increases the constant factors due to additional reweighting and residual computations.

A summary of the per-iteration complexities is provided in Table 1.

Table 1: Per-iteration computational complexity of the proposed clustering algorithms

| Algorithm | Per-Iteration Complexity | Dominant Operations |
| --- | --- | --- |
| Classical Crisp k-Means | $O(mnk)$ | Distance computation and centroid update |
| Classical Fuzzy c-Means | $O(mnk+nk^2)$ | Matrix multiplications and membership updates |
| Robust Crisp k-Means | $O(mnk)$ | Weighted distance computation and centroid update |
| Robust Fuzzy c-Means | $O(mnk+nk^2)$ | Weighted membership updates and centroid update |

# 5. Conclusion

In this note, we revisited classical clustering formulations from a matrix factorization perspective. While the equivalence of k-means clustering and matrix factorization has been noted in prior literature, we formalized and rederived the result in a rigorous manner, following the unpublished derivation by Bauckhage. As our first main contribution, we extended this equivalence to fuzzy c-means clustering—a result that, to the best of our knowledge, has not been explicitly shown before. This formulation allowed us to express both crisp and fuzzy clustering as matrix approximation problems, thereby providing a natural pathway to developing robust variants using the $\ell_{1,2}$-norm.

To this end, we introduced robust formulations for both clustering paradigms based on minimizing the sum of $\ell_2$-norms of the column residuals. These variants are particularly well-suited to handling sample-level corruption or outliers, which are common in high-dimensional or noisy real-world data. We derived alternating minimization algorithms for the Frobenius-norm versions, and IRLS-based solutions for the $\ell_{1,2}$-norm robust formulations. Theoretical convergence guarantees were provided for all variants.